\pdfoutput=1
\documentclass[11pt]{article}

\usepackage{graphicx}
\usepackage{fancyref}

\usepackage{EMNLP2023}

\usepackage{times}
\usepackage{latexsym}
\usepackage{tcolorbox}
\usepackage{float}

\usepackage[T1]{fontenc}
\usepackage[utf8]{inputenc}

\usepackage{microtype}

\usepackage{inconsolata}

\title{Osiris: A Lightweight Open-Source Hallucination Detection System}

\author{Alexander Shan \\
  Stanford University \\
  Dpt of Computer Science \\
  \texttt{azshan@stanford.edu} \\\And
  John Bauer \\
  Stanford HAI \\
  \texttt{horatio@stanford.edu} \\\And
  Chris Manning \\
  Stanford University \\
  Dpt of Computer Science \\
  \texttt{manning@stanford.edu}}

\begin{document}
\maketitle
\begin{abstract}

Retrieval-Augmented Generation (RAG) systems have gained widespread adoption by application builders because they leverage sources of truth to enable Large Language Models (LLMs) to generate more factually sound responses. However, hallucinations—instances of LLM responses that are unfaithful to the provided context—often prevent these systems from being deployed in production environments. Current hallucination detection methods typically involve human evaluation or the use of closed-source models to review RAG system outputs for hallucinations. Both human evaluators and closed-source models suffer from scaling issues due to their high costs and slow inference speeds. In this work, we introduce a perturbed multi-hop QA dataset with induced hallucinations. Via supervised fine-tuning on our dataset, we achieve better recall with a 7B model than GPT-4o on the RAGTruth hallucination detection benchmark and offer competitive performance on precision and accuracy, all while using a fraction of the parameters. Code is released at our repository.\footnote{\url{https://github.com/JudgmentLabs/osiris-detection}}

\end{abstract}

\section{Introduction}
    Hallucination detection in large language models (LLMs) is a critical
challenge in ensuring the reliability of AI-generated text, particularly in retrieval-augmented generation (RAG) systems \cite{xu2024hallucination} \cite{mckenna2023sources} \cite{banerjee2024llms}. Despite LLMs achieving remarkable performance in tasks like summarization, question-answering, and sentiment analysis, they frequently generate hallucinated responses (statements unsupported by context documents). This issue poses significant risk in high-stakes applications like healthcare, finance, and law, where misinformation can have severe consequences.

    Existing approaches to hallucination detection in RAG systems remain
inadequate. While retrieval mechanisms like semantic search \cite{sawarkar2024blended} \cite{purwar2023keyword}, embedding-based retrievers \cite{reichman2024dense} \cite{bhattarai2024heal} \cite{rau2024context}, and ranking enhancements improve context relevance, they do not prevent LLMs from contradicting retrieved evidence. Recent techniques, including Chain-of-Thought prompting \cite{wei2022chain}, post-training refinements like Direct Preference Optimization \cite{song2024rag}, and test-time interpretability methods \cite{sun2024redeep}, have helped mitigate hallucinations. However, these solutions still struggle with multi-hop reasoning, where models must synthesize information across multiple documents to determine factuality. Current hallucination evaluation models, including GPT-4o-based supervision, rely on distilled datasets that focus on single-hop question-answering and fail to generalize to real-world RAG settings.

    To address these limitations, we introduce Osiris-7B, a model optimized
for hallucination detection in multi-hop RAG contexts. Fine-tuned on multi-document question-answering tasks, Osiris-7B surpasses GPT-4o in recall, making it particularly useful in industry applications where identifying hallucinations is of utmost importance. By prioritizing recall over precision, Osiris-7B ensures that human reviewers can focus on flagged responses rather than manually verifying every model output. This significantly reduces the burden of exhaustive review workflows, while still maintaining high reliability.

    Empirical results demonstrate that Osiris-7B outperforms GPT-4o in
hallucination detection, improving recall by 22.8\% on the RAGTruth benchmark while maintaining competitive precision and F1 scores. Additionally, Osiris-7B achieves faster inference speeds (141.98 tokens/s vs. GPT-4o’s 97 tokens/s) \cite{kwon2023efficient} \cite{gpt4o2024}, making it a practical solution for real-time hallucination detection in industry settings.

Our contributions include:
\begin{itemize}
    \item \textbf{Osiris-7B} achieves higher recall than GPT-4o by 22.8\% in hallucination detection for the RAGTruth benchmark
    \item Significant improvements over base models through fine-tuning on the Qwen 2.5 series, with average increases of 32.98\% in recall, 4.55\% in precision, and 17.98\% in F1 scores
    \item Faster inference speed than closed-source alternatives while requiring less computational resources
\end{itemize}

\begin{figure}[H]
    \centering
    \includegraphics[width=\columnwidth]{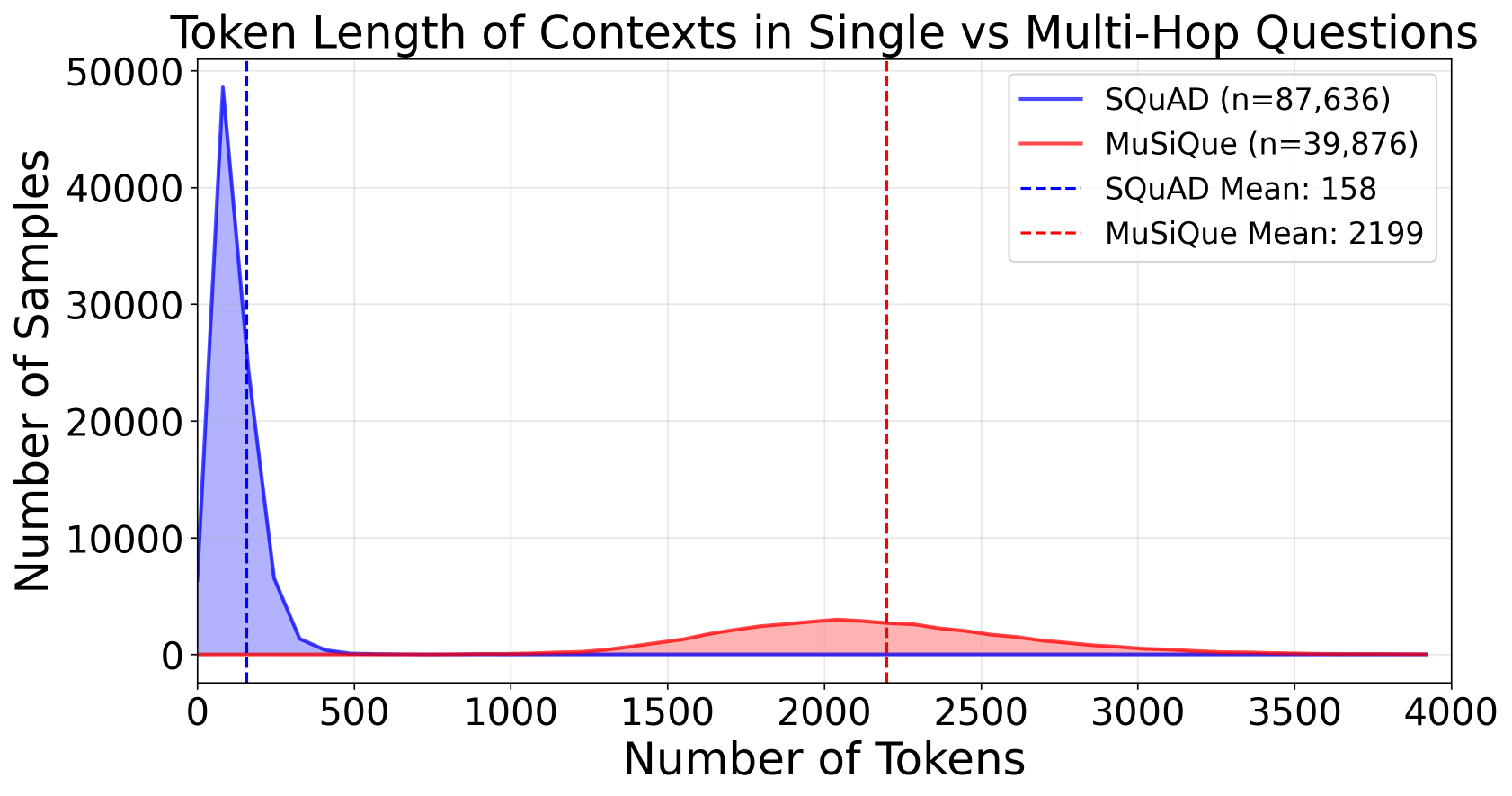}
    \caption{Distribution of Token Context Lengths}
    \label{fig:token_distribution}
\end{figure}

\section{Methodology}

To enhance hallucination detection performance, we built a data perturbation pipeline to construct fine-tuning data for developing Osiris-7B. For dataset construction, we began by selecting multi-hop QAs, as these require models to navigate through more diverse contexts. We then perturbed both positive and negative examples through our data pipeline to distill capabilities for detecting hallucinations. The following sections will address these methods in detail.

\subsection{Dataset}

\paragraph{Source} The MuSiQue \cite{TBKS2021} dataset is comprised of multi-hop reasoning questions, constructed by collecting single-hop questions from Wikipedia-based datasets such as SQuAD \cite{rajpurkar2016squad}, Natural Questions \cite{kwiatkowski2019natural}, MLQA \cite{lewis2019mlqa}, T-Rex \cite{elsahar2018t}, and Zero Shot RE \cite{levy2017zero}. A single-hop question is one that can be answered by retrieving information from a single document or source. These questions are typically very direct, requiring only the identification of the specific line containing the answer. An example from SQuAD is shown in Appendix \ref{appendix:single_hop}.
In contrast, multi-hop questions require synthesizing information from multiple documents or sources to arrive at an answer. These questions demand more reasoning steps, as the model needs to examine entire sources and piece together information to formulate a complete response. An example from MuSiQUe is shown in Appendix \ref{appendix:multi_hop}.

MuSiQue creates complex multi-hop questions by concatenating related single-hop questions, requiring sophisticated reasoning across multiple documents from diverse sources. These questions span 2-4 reasoning hops, significantly more complex than traditional single-hop RAG questions, and necessitate processing approximately 20 contextual paragraphs on average. Figure \ref{fig:token_distribution} compares token distributions between single-hop datasets like SQuAD and multi-hop datasets like MuSiQue, highlighting the substantial contextual information required. Working with these larger contexts demands higher levels of processing and more complex reasoning steps to properly address the questions.

\begin{figure*}[htbp]
    \centering
    \includegraphics[width=\textwidth]{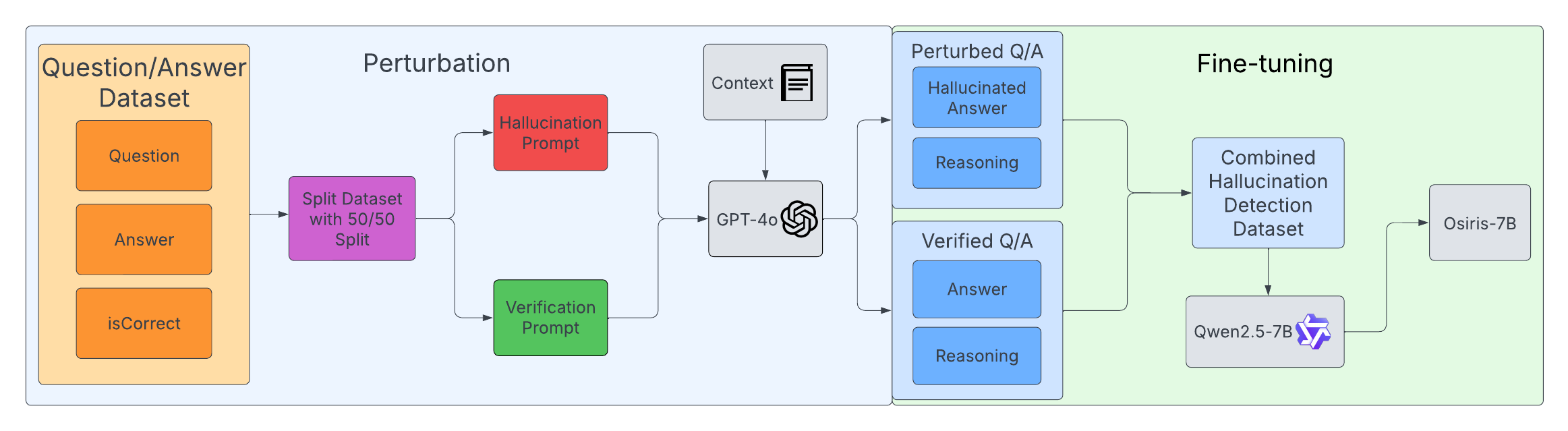}
    \caption{Q/A Dataset Perturbation Pipeline}
    \label{fig:perturb_pipeline}
\end{figure*}

\paragraph{Structured Reasoning for Hallucination Detection}  
Structured multi-hop reasoning frameworks provide a strong foundation for training hallucination detection models by enforcing explicit evidence retrieval and integration across multiple reasoning steps. Unlike single-hop QA datasets, which allow models to extract answers directly from the context, multi-hop QA datasets necessitate a sequential reasoning process, reducing the likelihood of models relying on spurious correlations or statistical artifacts. This property is particularly advantageous for hallucination detection, as it enables models to distinguish between answers that are genuinely supported by retrieved evidence and those that merely align with common knowledge or partial truths.  

\paragraph{Limitations of Traditional Verification}  

Traditional verification mechanisms often struggle in cases where hallucinated responses appear plausible despite lacking explicit support. However, training on multi-hop QA datasets conditions hallucination detection models to expect and enforce strict interdependencies between reasoning steps, improving their ability to identify responses that deviate from a well-grounded evidential chain \cite{10447488}. By requiring stepwise integration of evidence, multi-hop frameworks enhance a model’s ability to verify factual claims, reducing instances where models confidently generate hallucinated responses that lack sufficient justification 
\cite{Huang_2025}.

\paragraph{Shortcut-Driven Reasoning}  
A key advantage of using MuSiQue over other multi-hop QA datasets for training hallucination detection lies in its mitigation of shortcut-driven reasoning by introducing unanswerable questions \cite{TBKS2021}.

Many QA models, even those trained on multi-hop datasets, exhibit a tendency to rely on statistical co-occurrences rather than explicit multi-hop reasoning, making them susceptible to generating or failing to detect fabricated responses that resemble correct ones \cite{TBKS2021}; \cite{shao2023enhancingretrievalaugmentedlargelanguage}. MuSiQue mitigates this by enforcing compositional question structures that demand genuine multi-hop reasoning rather than relying on shallow heuristics.

This phenomenon has been extensively studied in the context of compositional reasoning, where many questions do not necessarily require true multi-hop inference to be answered correctly \cite{min2019compositionalquestionsnecessitatemultihop}. Furthermore, adversarial evaluation and training approaches have been developed to counteract reasoning shortcuts and improve robustness in multi-hop QA \cite{jiang2019avoidingreasoningshortcutsadversarial}.

By enforcing stepwise reasoning, well-made multi-hop QA datasets like MuSiQue ensure hallucination detectors verify whether each inference step is explicitly supported by evidence. This structured training strengthens their ability to distinguish well-supported conclusions from seemingly reasonable but unsubstantiated claims. Unlike traditional QA datasets, which may reinforce reliance on surface-level cues, multi-hop QA datasets compel models to track information dependencies, improving robustness against hallucination.

\paragraph{Hard Distractors and Contrastive Questions}  
Multi-hop question answering (QA) datasets that incorporate hard distractors and contrastive unanswerable questions further enhance their suitability for hallucination detector training. A common failure mode in hallucination detection arises when models struggle to differentiate between misleading yet contextually plausible distractors and genuinely verifiable claims. By exposing models to adversarial distractors designed to resemble supporting facts, multi-hop QA datasets provide a challenging training environment that teaches hallucination detectors to scrutinize context more rigorously.

In multi-hop question answering, adding hard distractors, which are misleading passages or reasoning chains, significantly tests model robustness. Recent research shows that even advanced language models suffer a steep performance drop when faced with plausible but incorrect supporting information \cite{Bhuiya2024AttentiveReaders}. In one study, state-of-the-art language models saw up to a 45 percent relative decrease in F1 score on a multi-hop QA task when presented with highly convincing distractor evidence \cite{Bhuiya2024AttentiveReaders}. While these models often ignore obvious lexical traps, they struggle with misleading reasoning paths, which remains a significant challenge \cite{Bhuiya2024AttentiveReaders}. This underscores the need for training QA systems to distinguish misleading from verifiable claims effectively.

Incorporating contrastive question pairs, which are questions that are similar except that one is answerable from the given data and the other is unanswerable, has proven effective for training models to recognize when a question lacks a valid answer. The MuSiQue-Full dataset was constructed with additional unanswerable contrast questions to enforce stringent multi-hop reasoning, forcing models to verify each hop of reasoning \cite{TBKS2021}. Similarly, a span-level contrastive learning approach explicitly trains models by pairing each answerable question with a nearly identical unanswerable counterpart \cite{Ji2022SpanCL}. This method boosted benchmark performance, yielding a 2.14-point absolute improvement in exact-match accuracy \cite{Ji2022SpanCL}. These results indicate that contrastive question training helps models develop a sharper sense of what information is missing or changed, improving their ability to detect unanswerable queries.

\begin{table*}[h]
\centering
\resizebox{\textwidth}{!}{%
\begin{tabular}{l|p{5cm}|p{4cm}|p{3cm}|p{2.5cm}|p{6cm}}
\hline
\textbf{Type} & \textbf{Context} & \textbf{Question} & \textbf{Original Answer} & \textbf{Perturbed Answer} & \textbf{Explanation} \\
\hline
Perturbed Q/A & \"Creeping Death\" is a song by the American heavy metal band Metallica ... the Scorpions had an artistic collaboration with the Berlin Philharmonic ... Metallica's similar collaboration (S\&M) with the San Francisco Symphony & "Who did the band of the song Creeping Death collaborate with?" & San Francisco Symphony & Berlin Philharmonic & The perturbed answer replaces 'San Francisco Symphony' with 'Berlin Philharmonic', another orchestra mentioned in the evidence, but not the one Metallica collaborated with for 'Creeping Death'. \\
\hline
Verified Q/A & In 1568, Spanish navigator Álvaro de Mendaña was the first European to sail through the archipelago, sighting the island of Nui. ... Tepuka is an island eighteen kilometers west of Fongafale, in the northwest of Funafuti, the main atoll of the Oceanian nation of Tuvalu & Who discovered the country Tepuka is located in? & Álvaro de Mendaña & N/A & The evidence text states that in 1568, Spanish navigator Álvaro de Mendaña was the first European to sail through the archipelago, sighting the island of Nui during his expedition. This indicates that Álvaro de Mendaña was the first European to discover the region where Tepuka is located, which is part of the Tuvalu islands. Therefore, the answer is correct as Álvaro de Mendaña is credited with the discovery of the area encompassing Tepuka. \\
\hline
\end{tabular}%
}
\caption{Perturbed and Verified Question-Answer Examples from Dataset}
\label{tab:perturbed_verified_QA}
\end{table*}

Design choices like hard distractors and contrastive questions have significant implications for hallucination detection and fact verification. A common failure mode of language models is overconfidently answering questions even when no supporting facts are available, leading to hallucinations \cite{Deng2024Gotcha}. Training models to recognize unanswerable questions reduces unwarranted claims \cite{Deng2024Gotcha}. Multi-hop reasoning is critical for detecting hallucinations in generated text \cite{Lei2025FactCG}. Fact-checking models trained with multi-hop synthetic data, such as graph-based reasoning chains, outperformed even GPT-4 in identifying factual errors \cite{Lei2025FactCG}. This highlights the necessity of enforcing stepwise reasoning to enhance factual accuracy and trustworthiness in natural language processing systems.

By reinforcing a disciplined, evidence-based reasoning approach, multi-hop QA datasets facilitate the development of hallucination detection models that are more adept at discerning valid multi-hop inferences from unsupported or fabricated claims, thereby advancing the reliability of factual verification in natural language processing systems.

\paragraph{Construction} We utilize an approach inspired by prior work, specifically Lynx \cite{RMMKQ2024} to construct a robust Question-Answer (QA) dataset aimed at training models to detect hallucinations produced by large language models. Our dataset leverages perturbing the MuSiQue QA dataset to induce hallucinated answers. Our method described in Figure~\ref{fig:perturb_pipeline} systematically addresses key challenges in training hallucination detection models by including verified truthful answers, deliberately misleading examples, and explicit reasoning justifying each answer's correctness or incorrectness.

Initially, contextually coherent Question/Answer pairs are extracted directly from MuSiQue. These pairs are subsequently verified using GPT-4o through a specialized verification prompt. detailed in the Appendix \ref{appendix:verification_prompt}, where the model explicitly provides reasoning that justifies the correctness of each answer based on the multi-hop context provided. This verification ensures the authenticity and reliability of positive examples in the dataset.

To effectively simulate realistic hallucination scenarios, we employ a dedicated hallucination prompt detailed in the Appendix \ref{appendix:halu_prompt} that encourages GPT-4o to generate plausible yet unsupported answers. These responses may contain information that appears in the context but is not actually supported by it, thereby increasing the complexity of hallucination detection. Crucially, GPT-4o provides explicit reasoning explaining why these hallucinated answers, despite their plausibility and potential contextual overlap, are not supported by the given context, aligning closely with the reasoning-centric approach described in Lynx.

\begin{table}[h]
    \centering
    \begin{tabular}{lc}
        \hline
        \textbf{Statistic} & \textbf{Value} \\
        \hline
        Total Samples & 39,876 \\
        Hallucinated Samples & 49.5\% \\
        Non-Hallucinated Samples & 50.5\% \\
        Average Token Context Length &  2199\\
        Average Token Reasoning Length &  45.5\\
        \hline
    \end{tabular}
    \caption{Perturbed MuSiQue Dataset Statistics}
    \label{tab:dataset_statistics}
\end{table}

\begin{table*}[h]
    \centering
    \begin{tabular}{lccc}
        \hline
        \textbf{Model} & \textbf{Recall} & \textbf{Precision} & \textbf{F1 Score} \\
        GPT-4o & 0.710 & 0.446 & 0.548 \\
        \hline
        \multicolumn{4}{c}{\textbf{Qwen2.5-Instruct Models}} \\
        \hline
        Qwen2.5-0.5B-Instruct & 0.098 & 0.250 & 0.140 \\
        Qwen2.5-1.5B-Instruct & 0.005 & 0.238 & 0.010 \\
        Qwen2.5-3B-Instruct   & 0.058 & 0.238 & 0.094 \\
        Qwen2.5-7B-Instruct   & 0.664 & 0.402 & 0.501 \\
        \hline
        \multicolumn{4}{c}{\textbf{Qwen2.5-Instruct + musique-v1.0}} \\
        \hline
        Qwen2.5-0.5B-Instruct + musique-v1.0 & 0.179 \(\uparrow\) & 0.270 \(\uparrow\) & 0.215 \(\uparrow\) \\
        Qwen2.5-1.5B-Instruct + musique-v1.0 & 0.170 \(\uparrow\) & 0.321 \(\uparrow\) & 0.222 \(\uparrow\) \\
        Qwen2.5-3B-Instruct + musique-v1.0   & 0.857 \(\uparrow\) & 0.353 \(\uparrow\) & 0.500 \(\uparrow\) \\
        \textbf{Qwen2.5-7B-Instruct + musique-v1.0} & \textbf{0.938} \(\uparrow\) & 0.366 \hphantom{\(\uparrow\)} & 0.527 \(\uparrow\) \\
        \hline
    \end{tabular}
     \caption{Performance metrics on the RAGTruth benchmark. Upward arrows (\(\uparrow\)) indicate improvements from base models, and bold values are better than GPT-4o.}
    \label{tab:results}
\end{table*}

\paragraph{Example} To concretely illustrate the distinctions between verified and hallucinated examples in our constructed dataset, Table~\ref{tab:perturbed_verified_QA} provides representative cases highlighting both types. The perturbed example demonstrates a subtle hallucination scenario: the original answer, "San Francisco Symphony," is intentionally changed to "Berlin Philharmonic," another orchestra mentioned within the provided context, but not one that collaborated with Metallica. Such perturbations create realistic yet unsupported claims, exemplifying the nuanced challenges faced by hallucination detection models.

Conversely, the verified example clearly demonstrates an accurate multi-hop inference explicitly confirming the the that the country Tepuka is located in (Tuvalu) was Álvaro de Mendaña. The model's reasoning shows that it cites the context via "sighting the island of Nui" and performs a multi-hop inference by connecting it to the fact that Tepuka is a part of the island nation Tuvalu. This carefully curated dataset structure enhances the model’s capability to differentiate factual from hallucinated content, supporting robust training and rigorous evaluation in practical industry scenarios. This is detailed in Table \ref{tab:dataset_statistics}.

\subsection{Evaluation}

We evaluated our models on the benchmark RAGTruth \cite{niu2023ragtruth}, a word-level hallucination corpus designed for various tasks within Retrieval-Augmented Generation (RAG). RAGTruth is a comprehensive benchmark consisting of responses generated by large language models (LLMs) to RAG questions, which have been manually annotated to ensure the highest standard of accuracy and reliability. For our evaluation, we utilized our fine-tuned models in conjunction with GPT-4o mini, employing a JSON prompt detailed in Appendix \ref{appendix:json} only when the output answers required correction, to assess performance on the RAGTruth dataset.

\subsection{Training Details}

We utilized the LLamaFactory \cite{zheng2024llamafactory} full fine-tuning script to fine-tune our dataset, musique-v1.0, on the Qwen 2.5 Family models \cite{yang2024qwen2}, resulting in the creation of Qwen-2.5-Instruct-musique-v1.0. Hyperparameters were chosen based on the Qwen 2.5 Family documentation. The comprehensive details of the fine-tuning process are presented in Appendix \ref{sec:training_details} and in Table \ref{tab:training_details}.

\section{Experiments}

 We fine-tuned the Qwen2.5 Instruct family models, resulting in Qwen2.5 Instruct-musique-v1.0.

\paragraph{RAGTruth} This fine-tuning process enhances the models' performance across key metrics, including recall, precision, and F1 score. Notably, the Qwen2.5-7B-Instruct + musique-v1.0 model achieves a remarkable improvement in recall, outperforming GPT-4o by \textbf{23.8\%} (0.938 vs. 0.710). In addition, musique-v1.0, demonstrates significant improvements in precision and F1 when fine-tuned on the Qwen Family Language models, as illustrated with more detail in Table \ref{tab:results}.

This enhancement in recall is achieved while maintaining competitive performance in precision and F1 score, highlighting the effectiveness of our approach. These results underscore the potential of musique-v1.0 to significantly boost the performance of language models in real-world applications, where high recall is crucial for minimizing overlooked errors.

\paragraph{Inference Speed}

We observed that inference speed is significantly faster on the smaller 7B model.
\textbf{Osiris-7B} is considerably smaller than closed-source models like GPT-4o and demonstrates much faster performance. Specifically, it achieves 141.98 tokens per second for input lengths of 6144 when using 4-bit quantization on a single A100 80GB via vLLM \cite{kwon2023efficient}, as compared to GPT-4o's 97 tokens per second \cite{gpt4o2024}.
This makes Osiris-7B much more scalable and faster, while also being able to run on significantly cheaper hardware, making it more suitable for real-time hallucination detection.

\section{Conclusion}

We developed Osiris-7B, a real-time hallucination detection model that achieves better recall than GPT-4o. By utilizing a multi-hop question answering dataset, we ensured that the model undergoes complex reasoning processes during fine-tuning, effectively identifying answers that are unsupported or contradicted by multiple evidence sources. With its high recall, Osiris-7B enables the industry to significantly reduce the number of evaluations required by humans, providing confidence with a recall that is 23.8\% higher on the RAGTruth benchmark. Due to its small size and high recall, we hope that Osiris-7B will be adopted by the industry and research community to explore more open-source and scalable methods for tackling the challenges of hallucination detection and evaluation.

\section*{Limitations}

While Osiris-7B demonstrates significant improvements in recall for hallucination detection, several limitations remain. The model struggles with industry-specific questions requiring specialized knowledge beyond its Wikipedia-sourced training data. Although recall improved substantially, precision and F1 scores did not scale proportionally, suggesting potential dataset limitations and higher false positive rates. In addition, instruction following in smaller models (0.5B, 1.5B, 3B) typically struggles with correct JSON output formatting, often necessitating GPT-4o mini for post-processing corrections. Future research should focus on enhancing dataset diversity with domain-specific data to improve precision and balance the recall-precision tradeoff.

\bibliography{anthology,custom}
\bibliographystyle{acl_natbib}

\clearpage
\appendix

\section{Data Examples}
\label{appendix:data}

\subsection{Single-Hop}
\label{appendix:single_hop}

\begin{minipage}{\textwidth}
Single-hop questions require the model to extract information directly from a single context passage without needing to combine information from multiple sources. These questions test basic reading comprehension and information retrieval capabilities. Below is an example from the SQuAD dataset that demonstrates this single-hop reasoning pattern.
\end{minipage}

\begin{tcolorbox}[colback=gray!5!white, colframe=gray!75!black, title=Single-Hop Question, width=\textwidth]
CONTEXT: "Traditionally, Switzerland avoids alliances that might entail military, political, or direct economic action and has been neutral since the end of its expansion in 1515. Its policy of neutrality was internationally recognised at the Congress of Vienna in 1815. Only in 2002 did Switzerland become a full member of the United Nations and it was the first state to join it by referendum. Switzerland maintains diplomatic relations with almost all countries and historically has served as an intermediary between other states. Switzerland is not a member of the European Union; the Swiss people have consistently rejected membership since the early 1990s. However, Switzerland does participate in the Schengen Area."\\

QUESTION: "When was Switzerland's policy of neutrality internationally recognized?\\

ANSWER: Congress of Vienna in 1815 write this text into a table for latex

\end{tcolorbox}

\clearpage
\subsection{Multi-Hop}
\label{appendix:multi_hop}

\begin{minipage}{\textwidth}
Multi-hop questions require models to stitch together pieces of information from multiple documents to answer a question. In this example from MuSiQue, the model must perform two distinct reasoning steps. First, it needs to identify that İsmail Keleş was born in Ankara by extracting information from one document. Then, using that answer, it must determine from a separate document that Melih Gökçek was the Metropolitan Mayor of Ankara at the time. This demonstrates the need for multi-step inference, as the question cannot be answered by considering either document in isolation. 
\end{minipage}

\begin{tcolorbox}[colback=gray!5!white, colframe=gray!75!black, title=Multi-hop Question, width=\textwidth]
RELEVANT CONTEXT:
Melih Gökçek has been the Metropolitan Mayor of Ankara since 1994 as a politician from the Welfare Party. He later joined the Virtue Party and then the AKP. Initially elected in the 1994 local elections, he was re-elected in 1999, 2004 and 2009. In the 2014 local election, Gökçek stood for a fifth term. The MHP metropolitan mayoral candidate for the 2009 local elections, conservative politician Mansur Yavaş, stood as the CHP candidate against Gökçek. In a heavily controversial election, Gökçek was declared the winner by just 1\% ahead of Yavaş amid allegations of systematic electoral fraud. With the Supreme Electoral Council and courts rejecting Yavaş's appeals, he has declared intention to take the irregularities to the European Court of Human Rights. Although Gökçek was inaugurated for a fifth term, most election observers believe that Yavaş was the winner of the election. \\ 

İsmail Keleş (born March 5, 1988 in Ankara, Turkey) is a Turkish sport shooter competing in the pistol events. By profession a non-commissioned officer at the Turkish Gendarmerie, the tall athlete at , is a member of Jandarma Gücü Sports Club, where he is coached by Muhammed Topal. \\ 

QUESTION:Who was in charge in the city where İsmail Keleş was born? \\ 

Q/A DECOMPOSITION: \\
Q: What is the place of birth of İsmail Keleş? \\
A: Ankara \\ 

Q: Who was in charge of Ankara? \\
A: Melih Gökçek

\end{tcolorbox}

\clearpage
\section{Prompts}
\label{appendix:prompts}

\subsection{Verification prompt}
\label{appendix:verification_prompt}

\begin{tcolorbox}[colback=blue!5!white, colframe=blue!75!black, title=Verification Prompt, width=\textwidth]
Given the question, evidence text, and an answer along with the truth value of the answer, you must give reasoning for why the truth value of the answer is correct. If the truth value of the answer is true, you must explain why the truth value of the answer is correct and cite the evidence text to support your answer. Otherwise, you must explain why the truth value of the answer is incorrect and, if possible, explain what the correct answer may be with citation. \\

\textbf{Question:} \{sample["question"]\} \\

\textbf{Evidence Text:} \{evidence\_text\} \\

\textbf{Answer:} \{sample["answer"]\} \\

\textbf{Format the output as:}
\begin{verbatim}
answer: <answer>
reasoning: <brief description of why the truth value is correct>
is_hallucinated: <truth value of the answer>
\end{verbatim}
\end{tcolorbox}

\subsection{Hallucination Prompt}
\label{appendix:halu_prompt}

\begin{tcolorbox}[colback=blue!5!white, colframe=blue!75!black, title=Hallucination Prompt, width=\textwidth]
Given the question, evidence text, and gold answer, generate a perturbed answer that is not supported by the evidence text. However, the perturbed answer may be in the evidence text, but not as the answer to the question. The difference between the perturbed answer and the gold answer should be subtle. No matter what, the perturbed answer should be different from the gold answer. \\

\textbf{Question:} \{sample["question"]\} \\

\textbf{Evidence Text:} \{evidence\_text\} \\

\textbf{Gold Answer:} \{sample["answer"]\} \\

\textbf{Format the output as:}
\begin{verbatim}
answer: <gold answer>
hallucinated_answer: <perturbed answer>
reasoning: <brief description of what was changed>
is_hallucinated: true
\end{verbatim}
\end{tcolorbox}

\clearpage
\subsection{JSON Fix Prompt}
\label{appendix:json}

\begin{tcolorbox}[colback=blue!5!white, colframe=blue!75!black, title=JSON Fix Prompt, width=\textwidth]

You are a JSON repair tool. Output only valid JSON, no explanations.
Common errors you fix:

1. Missing commas between array items: ["item1" "item2"] → ["item1", "item2"] \\

2. Unclosed brackets: \{"list": [\{"item": "value"\} → \{"list": [\{"item": "value"\}]\} \\

3. Missing quotes: \{list: [value]\} → \{"list": ["value"]\} \\

4. Trailing commas: ["item1", "item2",] → ["item1", "item2"] \\

5. Unstructured lists:  
"Hallucinations:  
1. First item 2. Second item"  
→ \{"hallucination\_list": ["First item", "Second item"]\} \\

6. Bullet points: • First item • Second item → \{"hallucination\_list": ["First item", "Second item"]\} \\

7. Numbered lists: "1) First item 2) Second item" → \{"hallucination\_list": ["First item", "Second item"]\} \\

8. Line-separated items: "First item \textbackslash n Second item" → \{"hallucination\_list": ["First item", "Second item"]\} \\

If you see a plain text response with phrases like "I found these hallucinations:" or "Hallucinated content:",
extract the listed items and format them as a proper JSON array in the hallucination\_list.

\begin{verbatim}
The JSON must follow this exact format:
{"hallucination\_list": ["span1", "span2"]}
or for no hallucinations: {"hallucination_list": []}

If you see {"type": "conflict", "span": "text"} format, extract ONLY the span value.
Example:
Input: {"hallucination_list": [
{"type": "conflict", "span": "text1"},
{"type": "baseless", "span": "text2"}
]}
Output: {"hallucination_list": ["text1", "text2"]}
\end{verbatim}
\end{tcolorbox}

\clearpage
\section{Training Details}
\label{sec:training_details}

\begin{minipage}{\textwidth}
Here are the hyperparameters used for fine-tuning, following the default Qwen2.5 documentation on LLaMA Factory:
\end{minipage}

\begin{center}
\begin{minipage}{\textwidth}
    \centering
    \begin{tabular}{lc}
        \hline
        \textbf{Parameter} & \textbf{Value} \\
        \hline
        Warmup Steps & 100 \\
        Weight Decay & 0.1 \\
        Per Device Train Batch Size & 4 \\
        Gradient Accumulation Steps & 4 \\
        DDP Timeout & 9000 \\
        Learning Rate & 5e-6 \\
        LR Scheduler Type & Cosine \\
        Number of Train Epochs & 3 \\
        BF16 & Enabled \\
        GPUs & 8 A100s \\
        \hline
    \end{tabular}
    \captionof{table}{Training Details}
    \label{tab:training_details}
\end{minipage}
\end{center}

\end{document}